# Nationality, Race, and Ethnicity Biases in and Consequences of Detecting AI-Generated Self-Presentations

Haoran Chu[1*], Linjuan Rita Men[1], Sixiao Liu[2], Shupei Yuan[3], Yuan Sun[1]

This study builds on person perception and human artificial intelligence interaction (HAII) theories to investigate how content and source cues, specifically race, ethnicity, and nationality, affect judgments of AI-generated content in a high-stakes self-presentation context: college applications. Results of a pre-registered experiment with a nationally representative U.S. sample ($N$ = 644) show that content heuristics, such as linguistic style, played a dominant role in AI detection. Source heuristics, such as nationality, also emerged as a significant factor, with international students more likely to be perceived as using AI, especially when their statements included AI-sounding features. Interestingly, Asian and Hispanic applicants were more likely to be judged as AI users when labeled as domestic students, suggesting interactions between racial stereotypes and AI detection. AI attribution led to lower perceptions of personal statement quality and authenticity, as well as negative evaluations of the applicant's competence, sociability, morality, and future success.

*Keywords*: Human-AI interaction; AI-detection; race and ethnicity; nationality; stereotype

"**I did it to make my paper results look better. Nobody at my school taught us morals or values**." This statement, quoted during a keynote presentation at the recent annual conference on Neural Information Processing Systems (NeurIPS 2024), sparked widespread backlash when the speaker singled out a specific group of international students to illustrate the maluses of artificial intelligence (AI), such as plagiarism (NeurIPS Board and 2024 Organizing Committee, 2024). While these remarks perpetuated harmful stereotypes about individuals of certain nationalities, races, and ethnicities, they also raise profound and novel questions: who is perceived as more likely to (mis)use AI, and what are the consequences of such perceptions?

Indeed, recent advancements in AI have led to unprecedented improvements in its integration into our social and cultural infrastructure. Large language models (LLMs) like OpenAI's GPT-4, Meta's Llama-3.2, and Google's Gemini can now produce textual and graphic content that rivals, if not exceeds, the quality of an average human writer in terms of language and content (Jakesch et al., 2023; Vaswani et al., 2017). The rapid development of these models has also accelerated AI's proliferation in both professional and personal settings. Content creators are increasingly using AI models to draft and edit communication materials, while more programmers are leveraging AI to streamline their coding processes and convert human language into computer-readable code (Porter & Zingaro, 2024; Yue et al., 2024). However, the widespread adoption of AI-powered tools has raised concerns about their social, political, and cultural impacts (Dwivedi et al., 2023). For instance, due to AI's ability to generate coherent stories, malicious actors can exploit this technology to mass-produce misinformation or clickbait at extremely low costs (Xu et al., 2023). Plagiarism is another critical issue, as noted above. Users may question whether they can or should take credit for content generated by AI, while AI developers face legal concerns about whether training their models on human-created content constitutes copyright infringement (Dien, 2023).

Given the ethical and legal challenges surrounding the use of AI, such as misinformation, plagiarism, and the unauthorized use of copyrighted materials, an important question arises: how can we distinguish AI-generated content from human-created content (Jakesch et al., 2023)? With their vast training data and powerful algorithms, LLMs can produce content that is almost indistinguishable from human-generated work (Dalalah & Dalalah, 2023; Kobak et al., 2024). Much research and entrepreneurial effort has been directed toward "outsmarting" AI with AI, such as developing classifiers to differentiate between AI- and student-written assignments (Hoover, 2024). However, despite some claims of high accuracy, research generally shows that AI detection algorithms perform underwhelmingly (Kobak et al., 2024; Partadiredja et al., 2020).

[1] College of Journalism and Communications, University of Florida; [2] Department of Population Health Sciences, College of Medicine, University of Central Florida; [3] Department of Communication, Northern Illinois University

[*] Correspondence concerning this paper should be addressed to Haoran Chu, College of Journalism and Communications University of Florida, Gainesville, FL, 32608; Email: chrischuphd@gmail.com.

If machines struggle to differentiate between AI- and human-generated writing, can humans do better? After all, language use is deeply embedded in human experience, and studies show that humans can be effective judges of AI-generated creative content, such as stories in textual formats (Chu & Liu, 2024). Unfortunately, research on humans' ability to detect AI-generated text suggests otherwise. Human accuracy tends to hover around the 50% chance threshold (Jakesch et al., 2023), similar to or even worse than specialized AI-detection models. Even more concerning is that human judges often rely on flawed heuristics when deciding whether a piece of text was created by AI or by humans (Jakesch et al., 2023). For example, one study found that humans believed AI-generated text contained more grammatical errors, rare phrases, and long words. Ironically, human writing actually exhibited more of these characteristics than AI-generated text (Jakesch et al., 2023).

While contemporary LLMs seem to have the upper hand in the ongoing "cat-and-mouse" game of generating and detecting AI-created content, the need for accurate attribution between human and AI contributions remains critical in many fields. AI-detection tools, such as classifiers and databases, primarily focus on the features of textual content, like word choice and linguistic style. In other words, they tend to "judge a book by its content." However, as we all know, humans don't always operate this way and often "judge a book by its cover," both literally and metaphorically. Perceptions of others, though often stereotypical and inaccurate, frequently shape how people evaluate someone's actions, personality, and potential future performance (Fiske, 2012). For instance, the perceived credibility of an individual can strongly influence how others assess the authenticity and trustworthiness of content they produce (Kiousis, 2001). Similarly, job applicants often receive different treatment based on the racial or ethnic connotations of their names (Bertrand & Mullainathan, 2004). Therefore, just as people rely on flawed content heuristics to determine whether a text was written by AI, it's reasonable to believe they may also use flawed external heuristics, such as source cues, to make such judgments.

The use of flawed content heuristics is detrimental, as it creates confusion in content consumption experiences and undermines audience trust (Jakesch et al., 2023). Reliance on flawed source heuristics is even more harmful. As the incidents at NeurIPS demonstrate, associating nationalities, races, and ethnicities with the misuse of AI perpetuates harmful stereotypes that are not only offensive and hurtful to the targeted groups but also damages public perception of AI and related technologies. This, in turn, impedes their constructive adoption. Therefore, it is critical to evaluate whether and how source heuristics influence people's judgments of AI-generated content and, more importantly, to examine the consequences of such evaluations. To address such goals, this study investigates whether source information, including the race, ethnicity, or nationality of the supposed message creator, affects people's ability to detect AI-generated content in a self-presentation context, beyond the influence of content heuristics identified in previous research. We focus on self-presentation because materials like college application essays, cover letters, and online dating profiles are important tools for impression management and are closely linked to individuals' future opportunities (Ert et al., 2016; Ma et al., 2017). In the meantime, focusing on self-presentation aligns with the existing research on AI content detection, allowing our findings to be compared with previous studies (Jakesch et al., 2019, 2023; Ma et al., 2017). In addition, while Human-AI Interaction research has shown that people often perceive AI-generated content as credible, even compared to human-generated content (Jakesch et al., 2019; Sundar, 2020), it remains unclear whether people perceive those who use AI to create content as more or less trustworthy.

In summary, the current study thus aims to address two underexplored questions in human-AI interaction. First, we integrate social psychological theories and research to expand the existing literature on AI detection, which largely focuses on content features. Second, we investigate the downstream effects of these judgments on evaluations of the content and, more importantly, on perceptions of the AI users or non-users. In the following sections, we explore how source heuristics might influence AI detection. We then draw on social psychology research to assess how perceived AI use impacts perceptions of content quality, as well as the perceived competence, warmth, and future performance of the individuals involved..

## Source Heuristics for Detecting AI-Generated Self-Presentations

As discussed above, the first key question this research seeks to answer is: Who is perceived more likely to use AI? Recent PEW reports show that younger, more highly educated people in the U.S. are more likely to use AI in their daily work, and liberal-leaning Democrats are generally less skeptical of AI-generated information than their Republican counterparts (McClain, 2024). The classic diffusion of innovation theory may also provide a useful framework to characterize the early adopters of AI (Raman et al., 2023). For example, people who are more tech-savvy or open-minded may be more likely to integrate AI models, such as LLMs, into their work or daily routines, even delegating important tasks like impression management to AI agents. While there is no clear consensus on the profile of typical AI users, it is likely that many of their characteristics overlap with those of power users of other advanced technologies, such as programming tools and specialized software (Hancock et al., 2020).

Although there is some theoretical and empirical evidence about who might use AI, little research has explored who the general public believes is more likely to use this new technology. While the stereotypical image of a power user, young, educated, and perhaps "nerdy", might describe an AI user, public perception of AI use likely goes beyond a person's technical proficiency. Indeed, AI's capabilities have surpassed many earlier technological advances, blurring the lines between human and machine intelligence. Unsurprisingly, some people may find this power both *inspiring*, as it holds promise for the future of humanity, and *intimidating*, given the uncertainties and complexities surrounding AI's mechanisms (Dwivedi et al., 2023). Notably, while many people have heard of AI through media coverage, the majority, even in developed countries like the U.S., have not had firsthand experience interacting with AI, at least knowingly (McClain, 2024). This combination of AI's power and the uncertainty surrounding its use may cast a shadow over the public's perception of it.

Although it may be extreme to assume that the public is outright hostile toward AI or its users, it is plausible that people hold mixed feelings about delegating critical tasks to AI, especially when it happens without disclosure. While AI models may not be considered bearers of copyright in a legal or cultural sense, the public might view using AI as a ghostwriter with some skepticism, particularly related to authenticity (Yue et al., 2024). Indeed, college professors have expressed frustration over AI-like assignments, and Hollywood screenwriters have successfully protested against the use of AI in creative work (Willis, 2023). In self-presentation contexts, such as writing a cover letter for a job application, a personal statement for college admission, or even a dating profile, people are even more likely to expect genuine, authentic self-expression rather than ghostwriting by AI. This sentiment is evident in the backlash against Google's recent ad campaign promoting its LLM, Gemini, which featured a story created on behalf of a young girl aspiring to become an Olympic athlete (Elias, 2024). Therefore, beyond the perceived competence in technology use, the public may also question the ethicality of AI use, particularly in settings where true and authentic human expression is expected.

The complex interplay between perceived capabilities and intent when evaluating others' use of AI calls for a more structured theoretical approach to formulating hypotheses and research questions. We turn to decades of research on stereotype-based person perception for guidance. Susan Fiske's (2012) stereotype content model (SCM) provides a useful perspective for disentangling the influence of different person perceptions on people's judgments of others' AI use. The SCM posits that human evaluations of each other often boil down to two key dimensions: perceived warmth and perceived competence (Fiske et al., 2018). Perceived warmth relates to the intent of others, with malevolent intent leading to a colder image, while perceived competence is associated with the evaluation of others' capabilities. More recent research suggests that the warmth dimension can be further subdivided into sociability and morality, whereas likeable and moral others are perceived as warmer (Kervyn et al., 2015).

People tend to form stereotypes of different demographic, cultural, and social groups based on their experiences, and these stereotypes profoundly shape subsequent judgments and evaluations. For instance, research shows that professions like scientists and entrepreneurs are typically perceived as high in competence but low in warmth, whereas blue-collar workers, like cashiers and janitors, are seen as high in warmth but low in competence (Fiske et al., 2018). Some stereotypes, such as those about racial and ethnic groups, may be less benign. For example, Asians and Jews are often stereotyped as high in competence but low in warmth, placing them in the "envied" group within the SCM framework (Fiske & Dupree, 2014). As previously mentioned, using AI may signal both technical competence and ethical concerns. In the context of the SCM, AI users may likely be perceived as falling into the high-competence/low-warmth category. Correspondingly, source cues that suggest this stereotype may influence judgments about whether content is generated by AI or a human, especially in high-stakes situations, such as college admissions.

In this research, we focus on how race, ethnicity, and nationality affect judgments of AI involvement in college application materials, as these key source cues can shape stereotypical views (Fiske et al., 2018). Specifically, race and ethnicity have been shown to profoundly affect people's perceptions of social actors. Field studies demonstrate that racial and ethnic stereotypes can negatively impact evaluations, with job applicants who have minority-associated names being viewed less favorably than those with typical White names in the U.S. (Bertrand & Mullainathan, 2004; Kang et al., 2016). Nationality may also play a role in evaluations, especially when determining whether a college essay was AI-generated. International applicants, particularly those from non-English-speaking countries, may be perceived as less proficient in English, leading to lower competence ratings in writing. Additionally, research suggests that international students are often perceived as more likely to cheat in academic settings, as different educational systems may have varying definitions of plagiarism (Martin et al., 2011; Mohammadkarimi, 2023). Synthesizing the theories above, we propose that race, ethnicity, and nationality may differentially influence judgments about AI involvement in self-presentation materials. As there is insufficient empirical evidence to make a directional hypothesis regarding how these source cues specifically impact AI judgments, we pose the following research question:

**RQ1.** How do race, ethnicity, and nationality influence people's judgments of AI involvement in self-presentation materials?

## Consequences of Detecting AI-generated Self-Presentations

As noted earlier, people's judgments of others' AI use may be intertwined with their perceptions of the characteristics of the individuals being evaluated. Consequently, these perceptions, often stereotypical and inaccurate, can significantly influence how people view other aspects of the person. From a competence perspective, while AI users are often more experienced with new technologies and tend to be more highly educated, the act of delegating tasks to AI could be seen as lazy or even akin to plagiarism (Dwivedi et al., 2023; Hoover, 2024), potentially diminishing others' perceptions of the user's competence. From a warmth perspective, AI ghostwriting may create a barrier between the content creator and the audience, leading to perceptions that the content lacks authenticity and good intent (i.e., low warmth). As a result, reduced perceptions of warmth could lead to less favorable evaluations of the writing quality, the individual, and ultimately, projections of their future performance.

While suspecting AI's involvement in content generation may lead to negative evaluations of those claiming credit, HAII research suggests otherwise (Sundar, 2020). For example, studies have found that people often view AI-produced writing as highly credible due to its perceived objectivity compared to human-generated content (Jakesch et al., 2019). Similarly, linguistic analyses of AI-generated stories have shown that AI excels in creating linguistically competent and logically coherent material, sometimes surpassing human writers (Chu & Liu, 2024; Dalalah & Dalalah, 2023). Therefore, perceiving a piece of self-presentation material as AI-generated could lead to higher evaluations of its quality. However, as previously discussed, AI-generated content often lacks the human touch present in human-generated material, which may lead to doubts about its authenticity.

Similar debates may arise when addressing the issue from a social psychological perspective. According to the SCM (Fiske et al., 2018), perceived AI use might be associated with higher competence but lower warmth. Specifically, suspected AI use in self-presentation contexts, such as college application materials, could be seen as signaling immoral behavior (e.g., ghostwriting), damaging perceptions of the individual's morality. Similarly, attributing the writing of self-presentation materials to AI could make the person seem less competent, just as plagiarism is often viewed as a sign of incompetence (Strangfeld, 2019). However, such effects may be reversed as previous research show that the "nerdy" image often associated with tech users conveys high competence but low sociability (Grønhøj et al., 2024).

Finally, perceived AI use may also affect how people evaluate an individual's future performance, especially in contexts like college applications, where self-presentation is critical. Specifically, using AI regularly might signal competence, leading people to believe that a student who is proficient in AI could excel in college and their future career. Yet, relying on AI to write self-representation materials may also indicate lack of confidence or perceived inability to produce independent work. As a consequence, the ambiguity surrounding the perceived competence, morality, and likability of AI use makes it difficult to predict how these perceptions will affect evaluations of future performance. Given the novelty of this inquiry and the lack of established empirical evidence, we propose the following research questions:

**RQ2.** How does attribution to AI or human sources influence people's perceptions of the quality and authenticity of self-presentation materials, such as college application essays?

**RQ3.** How does attribution to AI or human sources for self-presentation materials influence people's perceptions of the individual's competence, morality, and sociability?

**RQ4.** How does attribution to AI or human sources for self-presentation materials influence people's beliefs about the individual's future performance?

### The Current Study

As outlined above, this study aims to address two under-explored questions in human-AI interaction research: (1) how source heuristics influence people's detection of AI- or human-generated content, and (2) how these judgments affect their perceptions of both the content and the individual. Like prior research on AI detection, we focus on self-presentation contexts, particularly college application materials, due to their relevance to person perception and the high stakes involved. We experimentally manipulated source cues, including the race, ethnicity, and nationality of the claimed author.

At the same time, content-related heuristics, such as linguistic style, have been shown to significantly affect people's judgments of AI-generated content (Jakesch et al., 2019). We incorporated these cues into our research design for two reasons. First, as people often use multiple cues when evaluating new information (Fiske et al., 2018; Kiousis, 2001), it is possible that content cues, such as the use of rare phrases and long words, might amplify the stereotypes people apply when assessing the source of self-presentation material. Second, content and source heuristics may interact to influence judgments, with linguistic features potentially reinforcing or mitigating the biases associated with source characteristics, such as race or nationality. Therefore, we propose our final research question:

**RQ5.** Do content heuristics interact with source cues to influence people's attribution of self-presentation content to human or AI?

### Method

#### Sample

Upon Institutional Review Board (IRB) approval (Approval Number: [Redacted for Review]), a pre-registered experiment was conducted in October 2024 (Registration: [Redacted for Review]). We utilized quota sampling to recruit a sample of participants with demographic characteristics representative of the U.S. public from Prolific, a crowdsourcing platform that connects researchers with diverse groups of participants. A total of 948 participants completed the survey and passed two attention-check questions. Of these, 644 participants also passed the two manipulation check questions, which asked them to recall the race, ethnicity, and nationality of the applicant featured in the application materials. It's important to note that these manipulation checks were administered after participants had answered all questions related to their assessment of the application material, the applicant, and their future outlook, to avoid sensitization. This delay may have inadvertently impaired their memory of the source heuristics, which were subtly manipulated, leading to the relatively high dropout rate. However, since biases and stereotypes related to race, ethnicity, and nationality often influence judgments implicitly, it is likely that the source heuristic manipulations still influenced participants' evaluations, despite imperfect recall. Nevertheless, the results reported below are based on the final sample of 644 participants, but the findings remain consistent with the larger sample.

The average age of participants in the final sample was 46.62 years (SD = 15.68). There were more women (340, 52.8%) compared to men (297, 46.1%) and participants identifying as another gender (7, 1.1%). In terms of race and ethnicity, 408 participants identified as non-Hispanic White (63.4%), followed by non-Hispanic Black or African American (80, 12.4%), Hispanic or Latino (61, 9.5%), Asian, Pacific Islander, or Native American (55, 8.5%), and other races/ethnicities (40, 6.2%). The median education level was a 4-year college degree, and the median household income was between US$50,000 and $74,999.

**Stimuli**

The experimental stimuli consisted of college application materials. Participants were instructed to imagine that they are serving as a college admission officer and review application materials, which included basic information about the applicant (e.g., name, intended major, high school) and an excerpt from the applicant's personal statement (see Figure 1 for an example). The applicant's last names were manipulated as a source heuristic indicating their race and ethnicity. Specifically, we randomly selected three names from the most common last names among non-Hispanic White, non-Hispanic Black or African American, Hispanic or Latino, and Asian communities in the U.S., based on census data (Word et al., 2008). In addition to name cues, we explicitly included the applicant's race and ethnicity in their background information. To avoid gender stereotypes, only the applicant's first initial was presented. A control condition was also included in which the applicant's race and ethnicity were not disclosed, and only the initial of their last name was shown. The applicant's nationality was manipulated both explicitly and implicitly, with international applicants labeled as "International Student (Visa Required)" and domestic applicants as "Domestic Student (No Visa Required)." To enhance the authenticity of the stimuli, applicants' high school names varied according to nationality. Domestic students were assigned "Smithville High School," while international students were assigned "United International School." In addition, the applicants' intended major was randomized to avoid category-based confounds. Majors included STEM (astrophysics), business, and humanities, with extracurricular activities adjusted to match the major. While the major significantly affected AI detection ($F(2, 641) = 8.06$, $p < 0.001$), it did not interact with the experimental factors to influence the primary outcome variables. Thus, we did not include it in the main analyses reported below. All other information remained consistent across the conditions.

Content heuristics were also experimentally manipulated. Specifically, we created two versions of each personal statement, incorporating common but flawed heuristics that people use to judge whether content is human- or AI-generated (Jakesch et al., 2023). In the "human-sounding" version, the personal statement included contractions (e.g., "I'm" instead of "I am"), recall of past events, first-person pronouns, and family-related words (e.g., "my parents inspired me …"). In the "AI-sounding" version, the heuristics included grammatical issues, rare bigrams, and long words (Jakesch et al., 2023). These heuristics were selected because they are context-appropriate (e.g., the use of swear words would not be appropriate for college applications) and were found to influence judgments, even though they are not actual features of AI-generated self-presentation materials (Jakesch et al., 2023). All personal statements were created by the research team in collaboration with AI models such as ChatGPT and Claude.

**Procedure and Materials**

After providing informed consent, participants were randomly assigned to evaluate one of 20 versions of the application materials (5 [race/ethnicity cue] x 2 [nationality cue] x 2 [content cue]). After reviewing the material, they were asked to indicate whether they believed the personal statement was written by the applicant or by AI, using a five-point scale (1 = "Definitely human-written" to 5 = "Definitely AI-generated") (Jakesch et al., 2023). We measured participants' perceived quality and authenticity of the personal statement using two sets of four semantic differential scales (e.g., quality: 1 = "Poorly written" to 5 = "Well-written"; authenticity: 1 = "Inauthentic" to 5 = "Authentic"). The scales were reliable (alpha = 0.86 for quality perception, and alpha = 0.94 for authenticity perception).

Participants' perceptions of the applicant were assessed using three scales from Fiske et al. (2018), measuring perceived competence (alpha = 0.93), sociability (alpha = 0.93), and morality (alpha = 0.94). Sample items included: "How competent (competence)/likable (sociability)/trustworthy (morality) do you think [applicant name] is?" Responses were recorded on a five-point scale (1 = "Not at all" to 5 = "A great deal"). We also created three additional sets of questions to assess participants' perceptions of the applicant's future performance, focusing on college admission potential (alpha = 0.89; e.g., "[Applicant name] will be admitted to a competitive university"), college performance (alpha = 0.90; e.g., "[Applicant name] will maintain a strong GPA in college"), and career performance (alpha = 0.91; e.g., "[Applicant name] will succeed in their future career"). Responses to these items were also measured on a five-point scale (1 = "Strongly disagree" to 5 = "Strongly agree").

| | |
|---|---|
| Name | C. Mueller |
| High School | Smithville High School |
| GPA | 3.8 |
| Intended Major | Business |
| Extracurriculars | Entrepreneurship Club, Internship at a Local Startup, Stock Market Simulation Club |
| Visa Status | Domestic Student (No Visa Required) |
| Identified Race/Ethnicity | Non-Hispanic White |

**C. Mueller's Personal Statement**

Ever since I was young, I've been fascinated by how businesses work. Watching my parents run our small store, I couldn't help but feel drawn to the world of commerce. I remember helping out on weekends, learning about customer service and managing inventory. This hands-on experience sparked my interest in studying business. I've always enjoyed working with others, and I'm excited about the opportunity to study business administration in college.

Figure 1. Stimuli sample

## Results

To address **RQ1**, which asks whether source cues such as the applicant's race, ethnicity, and nationality influence judgments of AI's involvement in writing their personal statement, we first conducted a two-way analysis of variance (ANOVA). The results indicated no significant effects for race/ethnicity ($F(4, 634) = 0.58, p = 0.68$, partial $\eta^2 = 0.004$), nationality ($F(1, 634) = 0.60, p = 0.44$, partial $\eta^2 = 0.001$), or their interaction ($F(4, 634) = 0.93, p = 0.44$, partial $\eta^2 = 0.006$) on AI detection. However, these findings do not rule out the possibility that source heuristics may interact with content heuristics to influence participants' source judgments (**RQ5**). To explore this, we conducted an additional ANOVA, including content cues as a predictor. The results showed that content cues significantly predicted AI detection ($F(1, 624) = 129.61, p < 0.001$, partial $\eta^2 = 0.172$), although the main effects of race/ethnicity ($F(4, 624) = 0.20, p = 0.939$, partial $\eta^2 = 0.001$) and nationality ($F(1, 624) = 0.94, p = 0.332$, partial $\eta^2 = 0.002$) remained non-significant. Notably, the three-way interaction between race/ethnicity, nationality, and content heuristics was a significant predictor of AI detection ($F(4, 624) = 2.69, p < 0.05$, partial $\eta^2 = 0.017$). Bonferroni-adjusted pairwise comparisons revealed that participants were more likely to believe that AI was involved in writing the personal statement for an international student than for a domestic student when the student's race was not disclosed and the personal statement included many AI-sounding heuristics, such as long words and rare phrases ($\Delta M = 0.59, p < 0.05$). Interestingly, the direction of this effect reversed when the personal statement contained more human-sounding cues ($\Delta M = -0.44, p = 0.09$), though this effect only approached statistical significance after adjusting for multiple comparisons.

To further investigate the significant three-way interaction, we conducted a regression analysis using the PROCESS macro in SPSS (Model 3) to explicate the interactive effects of source and content cues on source judgment (Hayes, 2017). The results, presented in Table 1, show that the dummy-coded race/ethnicity variables contrasting the control condition with Hispanic/Latino or Asian applicants interacted with both international status and content heuristics to influence source judgment. A closer inspection of the conditional effects revealed that the applicant's international status significantly affected source judgment only when race or ethnicity information was not disclosed and the personal statement included many AI-sounding heuristics ($b = 0.59$, $p < 0.05$), consistent with the ANOVA findings. Additionally, human-sounding personal statements from a domestic student were less likely to be judged as AI-generated when attributed to Hispanic/Latino applicant than when the race is not disclosed ($b = -0.51$, $p < 0.05$). Further examination of the conditional effects of content heuristics on source judgment (Table 2) revealed that participants were more likely to attribute AI-sounding personal statements to AI when the applicant was a non-Hispanic White, Black/African American, or racially ambiguous international student than when the students of these racial and ethnic groups were labeled as domestic students. Conversely, content heuristics had a stronger effect on source judgment when Asian or Hispanic applicants were labeled as domestic rather than international students.

**RQ2** asks whether source judgment influences participants' evaluation of the personal statement's quality and authenticity. We conducted two additional ANOVA models using source and content cues as predictors. The results indicated no significant effects of source cues. However, content heuristics significantly predicted both perceived quality ($F(1, 623) = 6.55$, $p < 0.05$, partial $\eta^2 = 0.010$) and authenticity ($F(1, 624) = 103.5$, $p < 0.001$, partial $\eta^2 = 0.14$). Human-sounding personal statements were perceived as having higher quality ($M = 3.91$, $SD = 0.75$) and authenticity ($M = 4.03$, $SD = 0.92$) than AI-sounding versions (quality: $M = 3.74$, $SD = 0.96$; authenticity: $M = 3.18$, $SD = 1.17$). Notably, the difference in authenticity perception was greater than the difference in quality perception. To further assess whether source judgment affected these perceptions, we conducted mediation analyses using the PROCESS macro (Model 12). As shown in Table 1, attributing a personal statement to AI led to lower perceptions of both quality and authenticity.

To address **RQ3** and **RQ4**, which ask whether source judgment affects perceptions of the applicant and their future performance, we ran additional ANOVA and regression models (Table 1). Content cues significantly predicted all person perception and future performance variables. Specifically, human-sounding personal statements led to more favorable evaluations of the applicant's competence ($F(1, 624) = 20.81$, $p < 0.001$, partial $\eta^2 = 0.032$), sociability ($F(1, 624) = 80.51$, $p < 0.001$, partial $\eta^2 = 0.114$), and morality ($F(1, 624) = 63.66$, $p < 0.001$, partial $\eta^2 = 0.093$). They were also associated with higher ratings for the likelihood of college admission ($F(1, 624) = 15.97$, $p < 0.001$, partial $\eta^2 = 0.025$), potential for excelling in college ($F(1, 624) = 21.45$, $p < 0.001$, partial $\eta^2 = 0.033$), and future career success ($F(1, 624) = 23.87$, $p < 0.001$, partial $\eta^2 = 0.037$). Similarly, source judgment significantly predicted all outcome measures, suggesting generally negative attitudes toward the use of AI in self-presentation (Table 1).

Of note, content and source cues interacted to influence participants' assessments of the applicant's college admission potential and expected college performance. Bonferroni-adjusted post hoc comparisons indicated that participants believed a domestic student (compared with an international student) who is Black ($\Delta M = 0.68$, $p < 0.01$) or racially ambiguous ($\Delta M = 0.83$, $p < 0.01$) was more likely to be admitted to college when their personal statement included more human-sounding cues. Similarly, participants believed that domestic students were more likely to be admitted to college than international students when race or ethnicity was not disclosed and the personal statement contained human-sounding cues ($\Delta M = 0.53$, $p < 0.05$). For college performance, a similar trend emerged. For international students who did not disclose their race or ethnicity, a human-sounding personal statement led to higher estimates of college performance ($\Delta M = 0.82$, $p < 0.001$). This effect was also significant for domestic students who were Black ($\Delta M = 0.55$, $p < 0.01$) or Hispanic ($\Delta M = 0.41$, $p < 0.05$). Additionally, Asian international students were believed to perform better in college than racially ambiguous international students when both submitted AI-sounding personal statements ($\Delta M = 0.69$, $p < 0.05$). These were the only significant pairwise differences after adjustment

## Discussion

This study sought to explore how content and source cues influence the detection of AI-generated content in self-presentation contexts, particularly focusing on college application materials. Our findings contribute to the growing body of literature on AI detection by revealing the relative importance of content and source cues, while also shedding light on how these judgments affect perceptions of the message, the person, and their projected performance.

Our results reaffirm the centrality of content cues in AI detection, aligning with existing research that emphasizes how textual features such as grammar,

Table 1

Results of the regression models predicting the key outcome variables

| | AI Detection | Perceived Quality | Perceived Authenticity | Perceived Competence | Perceived Sociability | Perceived Morality | College Admission Potential | College Performance | Career Success Potential |
|---|---|---|---|---|---|---|---|---|---|
| Intercept | 2.7 | 4.58 | 5.61 | 5.16 | 4.80 | 5.04 | 4.72 | 4.93 | 4.98 |
| Content Heuristics [1] | 0.3 | -0.17 | -0.35* | -0.2 | -0.37 | -0.27 | 0.18 | 0.06 | -0.08 |
| AI Detection | - | **-0.24***  | **-0.66***  | **-0.47***  | **-0.41***  | **-0.58***  | **-0.43***  | **-0.39***  | **-0.40***  |
| Non-Hispanic White [2] | -0.13 | -0.37 | -0.17 | **-0.41*** | -0.3 | -0.21 | -0.01 | -0.06 | 0.02 |
| Non-Hispanic Black [2] | -0.16 | 0.2 | 0.23 | 0.08 | 0.28 | 0.36 | 0.35 | 0.2 | 0.21 |
| Hispanic or Latino [2] | **-0.51*** | -0.14 | 0.00 | -0.15 | -0.03 | 0.06 | 0.13 | 0.15 | 0.1 |
| Asian [2] | -0.44 | 0.08 | 0.22 | 0.00 | -0.01 | 0.08 | 0.4 | **0.34*** | 0.23 |
| International Status [3] | -0.44 | -0.07 | 0.01 | -0.07 | 0.03 | 0.14 | 0.34 | 0.22 | 0.09 |
| Content[1] × White[2] | -0.02 | **0.66*** | 0.35 | **0.55*** | 0.35 | 0.35 | 0.12 | 0.32 | 0.22 |
| Content[1] × Black[2] | 0.21 | -0.25 | -0.19 | -0.2 | -0.40 | -0.27 | **-0.56*** | -0.34 | -0.23 |
| Content[1] × Hispanic[2] | **0.87**** | 0.47 | 0.16 | 0.4 | 0.02 | 0.14 | -0.06 | -0.02 | 0.2 |
| Content[1] × Asian[2] | **0.85**** | 0.02 | 0.07 | 0.47 | 0.2 | 0.31 | -0.03 | 0.01 | 0.31 |
| Content[1] × International [3] | 0.48 | 0.11 | 0.01 | 0.15 | -0.03 | 0.03 | -0.45 | -0.36 | 0.07 |
| International Status × White[2] | 0.40 | 0.20 | 0.10 | 0.13 | 0.02 | -0.08 | -0.26 | -0.05 | -0.02 |
| International Status × Black[2] | **0.87**** | -0.19 | -0.21 | -0.18 | -0.35 | -0.37 | -0.54 | -0.21 | -0.13 |
| International Status × Hispanic[2] | **0.90**** | 0.19 | -0.02 | 0.16 | -0.07 | -0.2 | -0.46 | -0.41 | -0.31 |
| International Status × Asian[2] | **1.03*** | -0.07 | -0.18 | -0.07 | -0.11 | -0.3 | -0.53 | -0.31 | -0.16 |
| Content[1] × International[3] × White[2] | -0.51 | -0.31 | -0.14 | -0.27 | -0.07 | -0.04 | 0.28 | 0.14 | -0.25 |
| Content[1] × International[3] × Black[2] | -0.47 | 0.49 | 0.46 | 0.54 | 0.8 | 0.65 | **1.10*** | **0.73*** | 0.4 |
| Content[1] × International[3] × Hispanic[2] | **-1.39**** | -0.51 | -0.12 | -0.35 | 0.13 | -0.03 | 0.47 | 0.44 | -0.05 |
| Content[1] × International[3] × Asian[2] | **-1.41**** | 0.26 | 0.25 | -0.06 | 0.07 | 0.04 | 0.7 | 0.61 | 0.02 |
| $R^2$ | 0.19 | 0.12 | 0.52 | 0.31 | 0.33 | 0.45 | 0.27 | 0.31 | 0.31 |

Note. *$p < .05$, **$p < .01$, ***$p < .001$; Content Heuristic = AI-sounding content features (1 = AI-sounding, 0 = human-sounding); [2] Non-Hispanic White, Non-Hispanic Black, Hispanic or Latino, Asian = Dummy-coded race/ethnicity variables comparing against "Not disclosed" [3] International Status = 1 for international students, 0 for domestic; significant effects are in boldface

Table 2

Conditional Effect of Content Heuristics[1] on AI Detection by Race and International Status

| Race/Ethnicity Group | International Status | Effects |
|---|---|---|
| Not Disclosed | Domestic | 0.36 |
|  | International | **1.32*** |
| Non-Hispanic White | Domestic | **0.78** |
|  | International | **1.30*** |
| Non-Hispanic Black | Domestic | **0.69** |
|  | International | **1.25*** |
| Hispanic or Latino | Domestic | **1.17*** |
|  | International | **0.81** |
| Asian | Domestic | **1.20*** |
|  | International | **0.82** |

*Note.* **$p < .01$, ***$p < 0.001$; significant effects are in boldface; [1] Content Heuristic = AI-sounding content features (1 = AI-sounding, 0 = human-sounding)

word choice, and linguistic patterns strongly influence judgments about AI involvement (Jakesch et al., 2023). Specifically, participants were more likely to attribute AI authorship to personal statements that exhibited what they perceived to be AI-sounding heuristics, such as the use of rare words or grammatical anomalies. These findings suggest that when evaluating written content, participants relied heavily on these familiar cues to make source judgments, consistent with earlier studies.

However, we also observed that nationality served as a significant, though more subtle, source cue in AI detection. In particular, participants were more likely to suspect AI involvement when the applicant was labeled as an international student, especially when the content contained AI-sounding cues. This aligns with existing stereotypes about international students being more likely to plagiarize or use external assistance (Martin et al., 2011; Mohammadkarimi, 2023), which may stem from broader concerns about academic integrity among non-native English speakers. Notably, Asian and Hispanic applicants were judged differently based on their nationality status than students of other racial or ethnic groups, with content cues having a greater impact when they were labeled as domestic students. This may reflect an underlying "foreigner" stereotype, where these groups are more readily associated with international status, and even with perceptions of dishonesty or incompetence.

Interestingly, no clear pattern emerged regarding the underlying drivers of AI detection judgments, but race, ethnicity, and nationality appeared to play a role in shaping participants' perceptions. While content cues were the dominant factor, it is difficult to disentangle whether participants' judgments were more driven by concerns about competence or integrity, particularly given the influence of stereotypes. The use of AI in self-presentation, especially when filtered through racial, ethnic, and national stereotypes, may signal a lack of competence, reinforcing pre-existing biases about certain groups being less or more independent and capable without the help of technologies such as AI. At the same time, AI use may also raise concerns about integrity, particularly in high-stakes settings like college applications, where the involvement of AI could be seen as dishonest or inauthentic, especially for marginalized groups often subjected to harsher scrutiny (Bertrand & Mullainathan, 2004). Stereotypes may amplify these concerns, with applicants from underrepresented or minority groups more likely to face assumptions of unethical behavior or incompetence (Fiske et al., 2018).

The downstream effects of AI detection on participants' perceptions of the message, the applicant, and their projected future performance were profound. Our results indicate that attributing a personal statement to AI led to significantly lower evaluations of both the quality and authenticity of the message. Human-sounding statements, by contrast, were rated as higher in quality and authenticity, underscoring the importance of perceived "human touch" in self-presentation contexts. This is consistent with prior research showing that AI-generated content, though often linguistically competent, lacks the creativity and grounded experience typically associated with human-generated content (Chu & Liu, 2024).

In terms of person perception, AI detection significantly diminished participants' evaluations of the applicant's competence, sociability, and morality. These results suggest that participants harbor negative attitudes toward individuals who are perceived to rely on AI for tasks that require personal expression, such as college applications. This aligns with the stereotype content model (Fiske, 2012), where individuals perceived as low in competence but low in warmth (or sociability and morality) are often viewed with suspicion or distrust. Additionally, these negative perceptions extended to assessments of the applicant's future performance. Applicants suspected of using AI were deemed less likely to succeed in college and their future careers, reflecting broader concerns about the appropriateness of AI in high-stakes contexts where personal effort and authenticity are valued.

While this study provides valuable insights, several limitations should be acknowledged. First, the use of an online sample from Prolific, which tends to include more tech-savvy participants, may have influenced the results. It is possible that participants more familiar with AI technology may be better equipped to detect AI-generated content or have different attitudes toward its use compared to the general population. Future studies should aim to replicate these findings with more diverse and less technology-oriented samples. Second, our sample size was limited by the inclusion of specific screening criteria, such as demographic quotas and manipulation checks. Although these steps were necessary to ensure the quality of the data, they reduced the number of participants in the final analysis. Future studies could benefit from larger and more diverse samples to increase the generalizability of the findings. Finally, our study relied on explicit manipulations of source cues, such as race, ethnicity, and nationality. While this approach was effective in identifying significant effects, it may not fully capture the implicit biases that participants bring to these judgments. Future research could explore more subtle or implicit manipulations of source cues, perhaps by incorporating more nuanced signals of identity or by allowing participants to form their own judgments without being explicitly informed of the applicant's background. Additionally, this study focused on one hypothetical college admission scenario to examine AI detection in self-presentation materials. Future research could investigate how AI detection and its broader effects function in other high-stakes contexts, such as job applications or online dating.

## Conclusion

In conclusion, this study highlights the complexities of detecting AI-generated content in self-presentation contexts and the significant impact of both content and source cues on these judgments. The findings suggest that while content heuristics are critical in AI detection, source cues like race, ethnicity, and nationality can amplify concerns about competence and integrity. Moreover, detecting AI use leads to unfavorable perceptions of both the message and the individual, with potential implications for their perceived future performance. As AI continues to integrate into more aspects of human life, understanding how it is perceived, especially in high-stakes settings, will remain an important area of inquiry.